\documentclass[10pt,twocolumn,letterpaper]{article}

\usepackage[pagenumbers]{cvpr} 
\usepackage{times}

\usepackage{amsmath,amsfonts,bm}

\def\eqref#1{equation~\ref{#1}}

\def\1{\bm{1}}

\DeclareMathAlphabet{\mathsfit}{\encodingdefault}{\sfdefault}{m}{sl}
\SetMathAlphabet{\mathsfit}{bold}{\encodingdefault}{\sfdefault}{bx}{n}

\usepackage{tabularx}
\usepackage{url}
\usepackage{booktabs}
\usepackage{diagbox}
\usepackage{makecell}
\usepackage{multirow}
\usepackage{minitoc}
\usepackage{colortbl}
\usepackage{comment}
\usepackage{tikz}

\usepackage{algorithm}
\usepackage{algorithmic}
\usepackage{amsmath}
\usepackage{todonotes}
\newcommand{\promptbubble}[1]{\todo[color=purple!20, inline]{#1}}

\definecolor{cvprblue}{rgb}{0.21,0.49,0.74}
\usepackage[pagebackref,breaklinks,colorlinks,allcolors=cvprblue]{hyperref}

\def\paperID{6413} 
\def\confName{CVPR}
\def\confYear{2025}

\title{Critic-V: VLM Critics Help Catch VLM Errors in Multimodal Reasoning}

\author{Di Zhang$^{1,2}$\thanks{These authors contributed equally.}, Jingdi Lei$^{2*}$, Junxian Li$^{3,2*}$, Xunzhi Wang$^{4,2*}$, Yujie Liu$^{5,2}$, Zonglin Yang$^{6,2}$, Jiatong Li$^{7,2}$ \\Weida Wang$^{8,2}$, Suorong Yang$^{9,2}$, Jianbo Wu$^{10}$, Peng Ye$^{11}$, Wanli Ouyang$^2$, Dongzhan Zhou$^2$\thanks{Corresponding author} \\
\small{ $^1$Fudan University},$^2$Shanghai Artificial Intelligence Laboratory, \small{$^3$Shanghai Jiaotong University, $^4$Nankai University,} \\ \small{ $^5$Shanghai University, $^6$Nanyang Technological University} \small{$^7$Hong Kong Polytechnic University, $^8$Tongji University, }\\ \small{ $^{9}$Nanjing University, $^{10}$University of California, Merced, $^{11}$Chinese University of Hong Kong}\\
\tt{\small{zhoudongzhan@pjlab.org.cn}} \\
\small{Paper and Dataset Link: \textcolor{blue}{https://huggingface.co/papers/2411.18203}} \\
\small{Code: \textcolor{blue}{https://github.com/kyrieLei/Critic-V}}
}

\begin{document}
\maketitle

\begin{abstract}
Vision-language models~(VLMs) have shown remarkable advancements in multimodal reasoning tasks. However, they still often generate inaccurate or irrelevant responses due to issues like hallucinated image understandings or unrefined reasoning paths. To address these challenges, we introduce Critic-V, a novel framework inspired by the Actor-Critic paradigm to boost the reasoning capability of VLMs. This framework decouples the reasoning process and critic process by integrating two independent components: the Reasoner, which generates reasoning paths based on visual and textual inputs, and the Critic, which provides constructive critique to refine these paths.
In this approach, the Reasoner generates reasoning responses according to text prompts, which can evolve iteratively as a policy based on feedback from the Critic. This interaction process was theoretically driven by a reinforcement learning framework where the Critic offers natural language critiques instead of scalar rewards, enabling more nuanced feedback to boost the Reasoner's capability on complex reasoning tasks. The Critic model is trained using Direct Preference Optimization (DPO), leveraging a preference dataset of critiques ranked by Rule-based Reward~(RBR) to enhance its critic capabilities. Evaluation results show that the Critic-V framework significantly outperforms existing methods, including GPT-4V, on 5 out of 8 benchmarks, especially regarding reasoning accuracy and efficiency. Combining a dynamic text-based policy for the Reasoner and constructive feedback from the preference-optimized Critic enables a more reliable and context-sensitive multimodal reasoning process. Our approach provides a promising solution to enhance the reliability of VLMs, improving their performance in real-world reasoning-heavy multimodal applications such as autonomous driving and embodied intelligence.
\end{abstract}

\section{Introduction}
\label{sec:intro}

\begin{figure}[!htbp]
    \centering
    \includegraphics[width=0.95\linewidth]{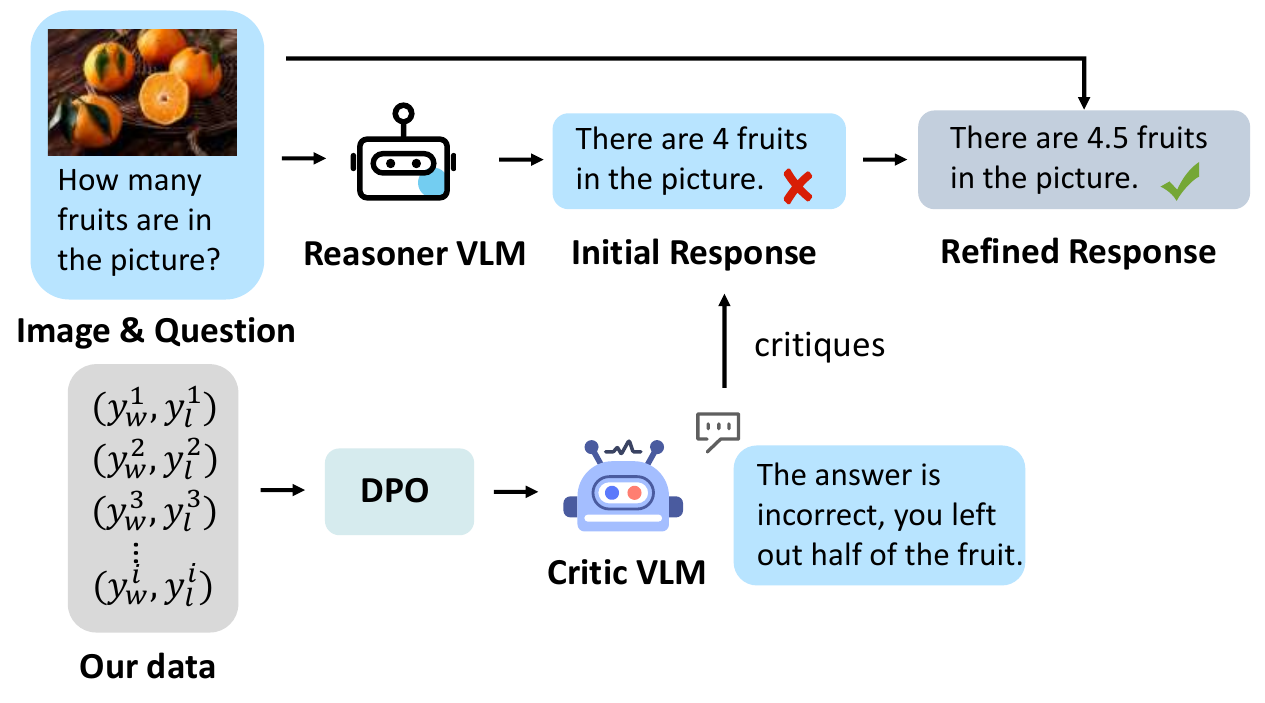}
    \caption{Offline training of critic model and response supervision for VLM. $y^i_w$ is preferred critique and $y^i_l$ is disfavored critique.}
    \label{fig:train}
\end{figure}
\noindent In recent years, Vision-Language Models (VLMs) have achieved significant advances in multimodal understanding and reasoning~\citep{GPT-4V, chen2024internvl, Qwen2VL, Llama-Vision, gemini-pro}. A major breakthrough has been the alignment of language and visual modalities, facilitated by techniques such as instruction tuning~\citep{zhu2023minigpt, alayrac2022flamingo, awadalla2023openflamingo}. This alignment allows VLMs to progress beyond basic image recognition, enabling them to perform dynamic content reasoning and handle complex question-answering based on visual inputs~\citep{yao2024minicpm, glm2024chatglm, internlmxcomposer2_5}. These advancements are pivotal for applications in embodied AI~\citep{jiang2022vima, driess2023palm} and autonomous driving~\citep{jiang2023vad, hu2022st}. Despite this progress, VLMs still encounter challenges, including a tendency to generate errors or irrelevant outputs that are unanchored in visual content~\citep{li2023evaluating, liu2023mitigating}. They may also over-rely on internal knowledge, sometimes neglecting the visual context~\citep{ziyang-etal-2024-vga}. Additionally, their sequential reasoning processes can lead to cascading errors, resulting in outputs that deviate from logical expectations~\citep{lightman2023let, bang2023multitask}.

Prior research primarily focuses on enhancing the intrinsic reasoning capabilities of VLMs through various strategies e.g. fine-tuning on curated datasets~\citep{zhang2024spa}, refining decoding methods~\citep{huang2024opera, woo2024don}, and test-time techniques like self-correction~\citep{he2024self}, self-consistency~\citep{dagan2024cast} and Self-Refine~\citep{madaan2024self} to address model flaws. Additionally, Silkie~\citep{li2023silkie} leverages direct preference optimization (DPO)~\citep{rafailov2024direct} to teach VLMs reasoning strategies using pairs of positive and negative samples. While these approaches have advanced the reasoning capabilities of VLMs, they often rely heavily on the model's internal abilities without incorporating external feedback, which may lead to erroneous or unreliable outputs. This raises a critical concern: How can we introduce high-quality supervision and feedback during the generation process of VLMs to effectively reduce errors and enhance the reliability of their reasoning path?

To address this concern, we introduce Critic-V, a novel framework based on reinforcement learning from human feedback (RLHF)~\citep{stiennon2020learning}. As shown in Figure~\ref{fig:train}, Critic-V features a Reasoner-Critic architecture, where the Reasoner generates reasoning paths based on visual content and related questions, while the Critic provides real-time feedback to refine these paths, enabling more accurate and dynamic reasoning, especially for complex tasks. 

However, the Critic evaluation capacity is still limited. To enhance the Critic's evaluative capacity, inspired by CriticGPT~\citep{mcaleese2024llm}, we introduce \textbf{V}ision \textbf{E}rror in\textbf{S}ertion \textbf{T}echnique~(VEST) which involves creating degraded versions of ground-truth VQA answers using GPT-4o\footnote{The version is chatgpt-4o-latest.}~\citep{OpenAI2024gpt4o} and obtaining critiques from multiple VLMs. The critic model is trained to assess these degraded answers, comparing them with the original ground truth. Additionally, we introduce a Rule-based Reward~(RBR) function using the Jaccard index to detect errors and reduce biases in feedback~\citep{mu2024rule}. 


Our experiments demonstrate that Critic-V significantly improves accuracy and reasoning efficiency compared to existing approaches like Self-Refine~\citep{madaan2024self}. These results underscore the importance of integrating an external, well-trained critic model into the reasoning process. Critic-V offers a promising solution for advancing the image understanding and reasoning capabilities of VLMs in real-world reasoning-heavy multimodal applications such as autonomous driving and embodied intelligence.

Our contributions can be summarized as follows:
\begin{itemize}

    \item \textbf{Integrated Reasoner-Critic Framework:} We propose a Reasoner-Critic framework that can integrate seamlessly with existing VLMs, significantly improving their performance in multimodal reasoning tasks by incorporating real-time feedback from an external Critic.

    \item \textbf{Large-Scale Multimodal Dataset:} We introduce a comprehensive dataset including 29,012 multimodal question-answer pairs with critiques generated by VEST and ranked by Rule-based Reward~(RBR). This resource can be used to enhance the Critic model, improving their ability to generate high-quality feedback.

    \item \textbf{Plug-and-play Critic model:} Our critic model can effectively guide VLMs in multimodal reasoning tasks while keenly identifying potential errors in the reasoning process. It provides proactive feedback on potential biases or errors rather than passively assess the quality of inference logic of VLMs, which enhances the overall multimodal reasoning capabilities of VLMs.

\end{itemize}

\section{Method}
\label{sec:method}

Multimodal reasoning remains a significant challenge for VLMs, which often struggle with inaccuracies when summarizing image content or addressing complex, reasoning-intensive questions. These unintentional errors can undermine the performance of VLMs in practical applications. To address this issue, we propose an approach inspired by the Actor-Critic framework, which separates the reasoning process from quality evaluation by incorporating two distinct, complementary modules: the Reasoner and the Critic.

The Reasoner is responsible for generating reasoning paths from both visual and textual inputs. Leveraging the principles of In-Context Reinforcement Learning (ICRL)~\cite{laskin2022context}, it uses prompt-based parameters to adapt its reasoning strategy during inference. By integrating visual content with textual descriptions, the Reasoner produces reasoning paths that are continuously evaluated and refined based on feedback from the Critic, enabling the model to improve the quality of responses particularly when faced with complex tasks.

The Critic functions as a quality evaluator for the reasoning paths generated by the Reasoner. By providing natural language feedback, the Critic generates gradient signals that guide the Reasoner in refining its strategy. This feedback loop encourages the Reasoner to minimize errors and enhance its reasoning capabilities over time, leading to more accurate and reliable outputs.

The following subsections provide a detailed discussion of the architecture and functionality of each module.

\begin{figure*}[!htbp]
    \centering
    \includegraphics[width=0.85\linewidth]{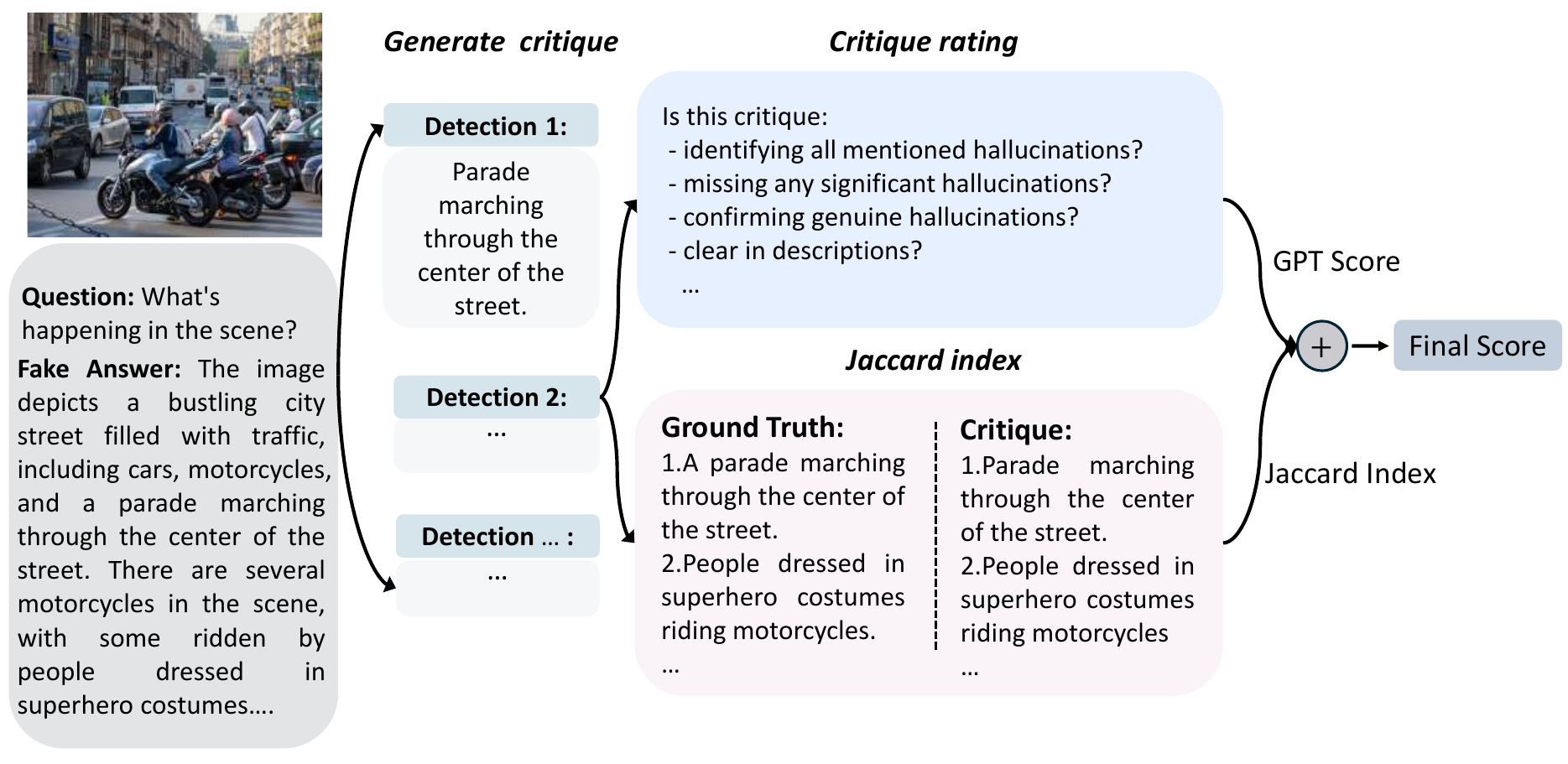}
    \caption{The scoring method combines GPT's evaluation with several predefined rules and the Jaccard index.}
    \label{fig:score}
\end{figure*}

\subsection{Reasoner}
To improve the reasoning process in reinforcement learning (RL), the Reasoner is responsible for generating reasoning actions \( a \) based on the current state \( s \), typically via a policy function \( \pi_{\theta^{\textit{reasoner}}}(a|s) \) parameterized by \( \theta^{\textit{reasoner}} \). The core goal is to optimize the reasoning strategy, often by adjusting these parameters through standard RL methods, such as policy gradient. As in policy gradient~\citep{sutton1999policy}, the update rule for the reasoner's parameters can be expressed as follows:
\begin{equation}
    \delta \theta^{\textit{reasoner}}_t = \nabla_{\theta^{\textit{reasoner}}_t} \log(\pi_{\theta^{\textit{reasoner}}_t}(a|s)) V(a|s),
\end{equation}

\noindent where \( V(a|s) \) represents the value function, which estimates the expected return for taking action \( a \) in state \( s \). This value function is typically parameterized by a critic model, which provides feedback that guides the updates to the reasoner’s policy.

However, as VLMs have become increasingly prominent in multimodal tasks, a challenge arises in adapting the traditional reinforcement learning framework to better handle these complex inputs. Specifically, rather than rely on a fixed parameterized policy, the reasoning process in VLMs can be driven by dynamic text prompts \( P^{\textit{reasoner}} \), which encapsulate the reasoning context and provide a more flexible approach to action generation. This shift allows for the integration of both visual and textual information, enabling the reasoning process to be guided by the context provided by the text prompt, instead of traditional policy parameters.

In this new approach, the reasoner’s policy update is no longer based solely on traditional parameterization but instead on the evolution of the text prompt. The update rule for the reasoner in this context can be described as follows:
\begin{equation}
    \delta \theta^{\textit{reasoner}}_t = \nabla_{\theta^{\textit{reasoner}}}\log\pi_{\theta^{\textit{reasoner}}}(P^{\textit{reasoner}}_t + \delta P^{\textit{reasoner}}_t, I) R_t,
\end{equation}

\noindent where \( P^{\textit{reasoner}}_t \) represents the current text prompt, \( \delta P^{\textit{reasoner}}_t \) is the critique (feedback) provided by the critic model, \( I \) is the input image, and \( R_t \) is the reward signal. This approach allows the reasoner to adaptively refine its actions through changes to the text prompt, which in turn leads to improved decision-making.

\begin{figure*}[!htbp]
    \centering
    \includegraphics[width=0.85\linewidth]{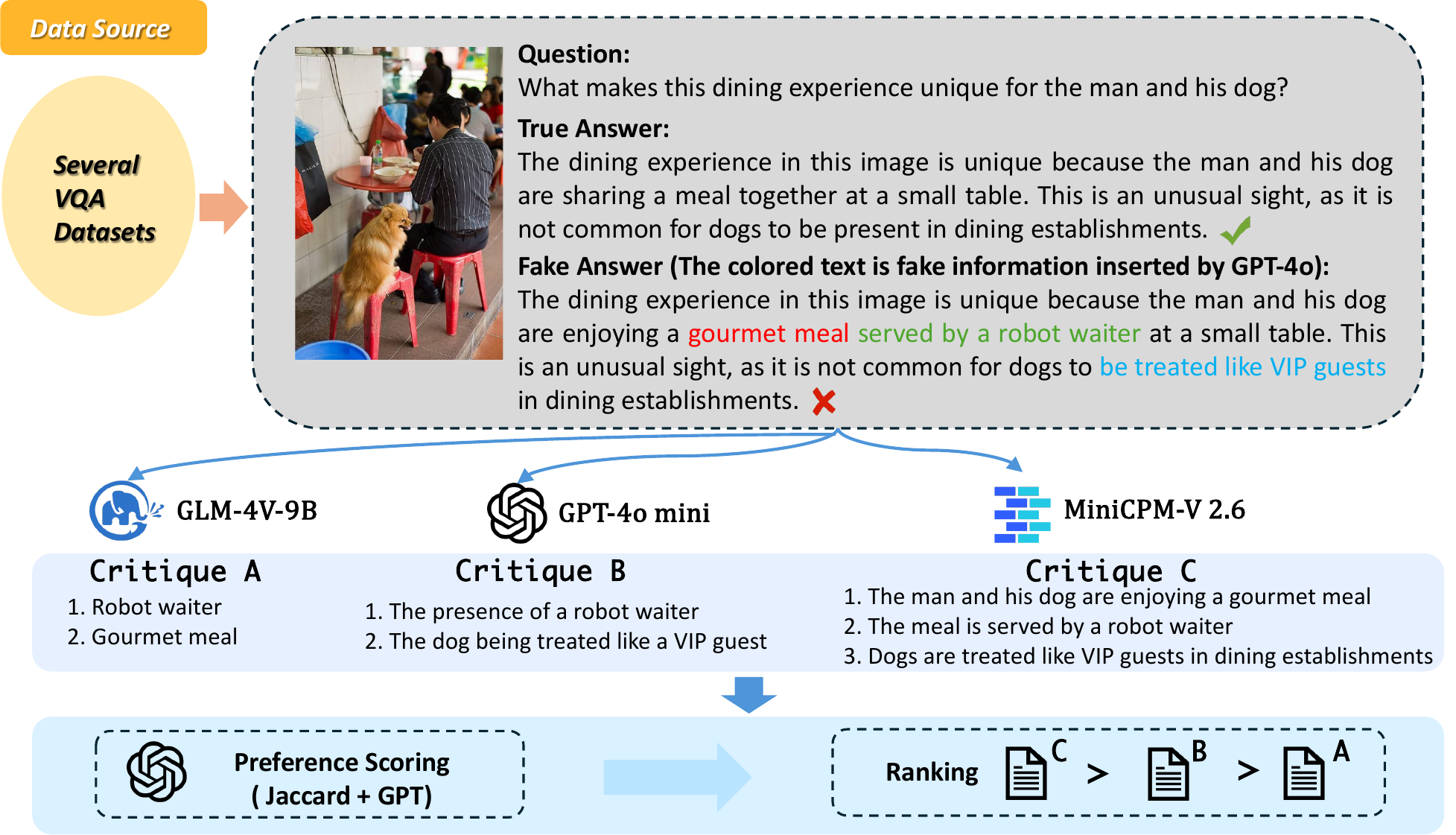}
    \caption{The annotation framework for our critique on the VisualQA~(critique-VQA) dataset. We collect questions and images from various sources, then use GPT-4o to generate a fake answer and employ three different VLMs to identify incorrect elements. Finally, we apply our proposed scoring method to calculate preference between different assessments.}
    \label{fig:data}
\end{figure*}

Despite the potential benefits of using text prompts, the challenge of computing stable and precise gradients for prompt updates remains. To address this, we leverage TextGrad~\citep{yuksekgonul2024textgrad}, a framework designed to provide a more intuitive and stable method for computing the gradients of text-based policies. TextGrad aims to improve the stability and accuracy of the text prompt update process, offering a more reliable alternative to traditional numerical gradient methods. The update rule for the text prompt, within the TextGrad framework, is given by:
\begin{equation}
    \delta P^{\textit{reasoner}}_t = \hat{\nabla}_{P^{\textit{reasoner}}_t}(\pi_{P^{\textit{reasoner}}_t}(a|s), V(a|s)),
\end{equation}
\noindent where \( \hat{\nabla} \) indicates the gradient computed using the TextGrad approach. This method significantly improves the robustness of prompt updates, ensuring that the reasoner can effectively learn from the feedback.

Nevertheless, while TextGrad improves stability, further refinement is still possible. To enhance the precision of gradient estimates, we introduce the critic model as an approximation to the optimal gradient. The critic model learns to predict the optimal updates for the text prompt by estimating the expected return of different actions in a given state. This approximation allows the reasoner to more effectively optimize its text prompts, guided by the Critic's feedback. The Critic’s role can be described as follows:
\begin{equation}
    \pi_{\theta^{\textit{critic}}}(\delta P^{\textit{reasoner}} | P^{\textit{reasoner}}) = \mathbb{E}[\pi_{\theta^{\textit{critic}}}(\delta P^{\textit{reasoner}} | P^{\textit{reasoner}}, s, a)].
\end{equation}
Finally, with the Critic’s guidance, the update rule for the text prompt becomes:
\begin{equation}
    P^{\textit{reasoner}}_{t+1} \leftarrow Update( P^{\textit{reasoner}}_{t}, \pi_{\theta^{\textit{critic}}}(\delta P^{\textit{reasoner}}), \eta),
\end{equation}
\noindent where $Update$ is a function that applies the Critic’s feedback to refine the text prompt \( P^{\textit{reasoner}} \), and \( \eta \) represents the learning rate, controlling the strength of each update.

\subsection{Critic Model}
In the Reasoner-Critic framework, the Critic serves a crucial role in providing evaluative feedback on the reasoning and generation processes of the model. Unlike traditional scalar rewards that assign a single numerical value, the Critic offers natural language feedback that is more nuanced and context-sensitive. This form of feedback is particularly valuable for complex tasks, as it enables the identification of subtle details in the reasoning process, including fine-grained errors, and logical inconsistencies. Scalar rewards, by contrast, often lack the depth needed for effective natural language reasoning, as highlighted in~\citep{golovneva2022roscoe}.

To update the Critic’s parameters, we start with a standard RL formulation, where the Critic’s policy is adjusted based on the feedback it provides to the reasoning model. The Critic’s policy is updated through the following equation:
\begin{equation}
\theta^{\textit{critic}}_{t+1} \leftarrow \theta^{\textit{critic}}_{t} + \eta \nabla_{\theta^{\textit{critic}}_t} \log(\pi_{\theta^{\textit{critic}}_t}(\delta P^{\textit{reasoner}}_t | P^{\textit{reasoner}}_t)) R_t,
\end{equation}
\noindent where \( P^{\textit{reasoner}}_t \) is the text prompt given to the VLM reasoner, and \( \delta P^{\textit{reasoner}}_t \) is the critique generated by the critic model. The term \( R_t \) represents the reward signal. 

To further enhance the Critic’s ability to generate more useful feedback, we thus shift from the scalar rewards fashion of policy gradient to preference-based training via DPO. Rather than optimizing a fixed reward, DPO focuses on training the Critic to distinguish between high-quality and low-quality critiques. This preference-based approach allows for a more subtle and context-aware form of learning, where the Critic improves by ranking critiques rather than directly optimizing for a scalar reward.

\begin{table*}[!tbp]
\centering
\renewcommand{\arraystretch}{1.5} 
\caption{Main results of VLMs on various benchmarks, reported as percentage scores. The \textbf{bolded} scores indicate the best performance on each benchmark. Additionally, we report the score improvements of Qwen2-VL-7B and DeepSeek-VL-7B compared to their original scores with the application of our method (+Critic-V).}
\label{tab:math-models-benchmark}
\resizebox{2.05\columnwidth}{!}{
\begin{tabular}{l|cccccc|cc}
\toprule
\hline
\multirow{2}{*}{Model} & \multicolumn{8}{c}{Benchmarks} \\
\cline{2-9}
 & RealWorldQA~\citep{realworldqa} & MMStar~\citep{chen2024we} & MMBench~\citep{liu2025mmbench} & SEEDBench~\citep{li2023seed} & ScienceQA~\citep{lu2022learn} & MMT-Bench~\citep{mmtbench} & MathVista~\citep{lu2024mathvista} & MathVerse~\citep{zhang2024mathverse}\\
\hline
Llama-3.2-11B-Vision~\citep{Llama-Vision}          & 57.8  & 49.8   & 65.8 & 62.2  & 67.8  & 47.9  & 48.6 & 24.31     \\
MiniCPM-V 2.6~\citep{yao2024minicpm}                 & 65.2  & 57.5   & 78.0 & 71.7  & 90.9  & 56.6  & 60.6 & 24.1     \\
InternVL2-8B~\citep{chen2024internvl}                  & 64.4  & \textbf{61.5}   & 79.4 & 76.2  & 89.2  & 54.8  & 58.3 & 30.3     \\
GPT-4V~\citep{yang2023dawn} & 61.4 & 57.1 & 74.3 & 71.6 & 81.4 & 55.5 & 49.9 & \textbf{54.4} \\
GeminiPro-Vision~\citep{team2023gemini} & 67.5 & 42.6 & 68.1 & 64.3 & 80.6 & 55.1 & 36.0 & 35.3 \\
LLaVA-v1.5-13B~\citep{liu2024improved} & 55.3 & 32.8 & 68.6 & 68.1 & 72.2 & 45.7 & 26.4 & 12.7 \\
ShareGPT4V-7B~\citep{chen2023sharegpt4v} & 56.9 & 33.0 & 69.5 & 69.4 & 69.4 & 45.1 & 25.7 & 17.4 \\
InternLM2-XC2~\citep{dong2024internlm} & 63.8 & 55.4 & 78.1 & 74.9 & \textbf{96.7} & 50.0 & 57.4 & 25.9 \\
\hline
Qwen2-VL-7B~\citep{Qwen2VL}           & 70.1  & 60.7   & 80.7 & 74.7 & 73.4(mm-only)  & 60.4   & 61.4 & 25.8     \\
\rowcolor{orange!20} 
\textbf{Qwen2-VL-7B+Critic-V}           & \textbf{74.9}(+4.8) & 56.2(-4.5)   & \textbf{82.8}(+2.1) & \textbf{76.5}(+1.8)  & 74.5(mm-only, +1.1)  & \textbf{62.0}(+1.6)   & \textbf{73.2}(+11.8) & 32.9(+7.1)  \\
\hline
DeepSeek-VL-7B~\citep{lu2024deepseek} & 58.1 & 37.1 & 73.5 & 70.2 & 61.7(mm-only) & 46.5 & 35.3 & 18.4 \\
\rowcolor{orange!20} 
\textbf{DeepSeek-VL-7B+Critic-V}         & 62.1(+4.0) & 41.4(+4.3)   & 79.0(+5.5) & 70.6(+0.4)  & 67.1(mm-only, +5.4)  & 53.6(+7.1)  & 53.1(+17.8) & 28.9(+10.5)  \\
\hline

LLaVA-v1.5-7B~\citep{liu2024improved} & 50.7 & 32.2 & 68.4 & 65.6 & 60.8 & 36.0 &37.8&26.0 \\
\rowcolor{orange!20} 
\textbf{LLaVA-v1.5-7B+Critic-V}           & 63.5(+12.8) & 38.4(+6.2)   & 73.8(+5.4) & 70.1(+4.5)  & 65.2(+4.4) & 47.4(+11.4)  & 53.1(+15.3) &30.5(+4.5) \\
\hline

\hline
\bottomrule
\end{tabular}}
\end{table*}

To generate preference data for training the Critic with DPO, we apply vision error insertion technique~(VEST) to question-image pairs from VQA datasets which is depicted in Figure~\ref{fig:data}. For each question-image pair, we use GPT-4o to insert one to five fake details into the answer. These fake details are erroneous and can simulate imperfections or errors in the reasoning or modality understanding process, creating a ground truth for the evaluation of the critique's quality. Several VLMs, including GLM-4V-9B \citep{glm2024chatglm}, GPT-4o mini \citep{gpt4omini}, and MiniCPM-V~\citep{yao2024minicpm}, are instructed to generate critiques that identify inaccuracies and highlight the weaknesses within the answers.

Then, we leverage a Rule-based Reward~(RBR)~\citep{mu2024rule} mechanism to evaluate the quality of critiques to construct the preference relationship. This reward mechanism evaluates the critique's statements by considering coverage, accuracy, and precision, particularly in identifying and addressing errors. Specifically, we use a scoring method to evaluate critiques based on how effectively they identify and describe inaccuracies. However, since longer critiques are more likely to contain extraneous information or “nitpicks” \citep{mcaleese2024llm}, we also incorporate the Jaccard index to adjust for potential bias towards false positives. As shown in Figure~\ref{fig:score}, the Jaccard index compares the set of errors inserted by GPT-4o (\(G\)) with the set of errors detected by the VLM (\(C\)) as follows:
\begin{equation}
Jaccard(G, C) = \frac{|G \cap C|}{|G \cup C|} = \frac{|G \cap C|}{|G| + |C| - |G \cap C|}.
\end{equation}
The final score for a critique is a combination of both the Jaccard index and the GPT-based score, where the GPT-based score serves as a regularization term in the scoring function:
\begin{equation}
Score(i) = Jaccard(i) + \alpha \times GPT(i), 
\end{equation} where $\alpha$ is hyperparameter to control the impact of GPT-4o's score on the final score~(the setting can refer to Appendix~\ref{appendix:dpo_hyp}). These preference scores allow us to rank various critiques based on their quality, which we then use to construct the critique-VQA dataset. This dataset consists of pairs of critiques with associated preference scores, providing the necessary data for training the critic model. Once the preference dataset is constructed, we proceed to apply DPO to optimize the base model, Qwen2-VL-7B~\citep{Qwen2VL}, thereby enhancing its ability to deliver more accurate and context-sensitive critiques. The dataset \( \mathcal{D}_{cri} = \{(Q^{(i)}, I^{(i)}, C_{w}^{(i)}, C_{l}^{(i)})\}_{i=1}^N \) consists of input questions \( Q^{(i)} \), corresponding images \( I^{(i)} \), the preferred critique \( C_{w}^{(i)} \), and the disfavored critique \( C_{l}^{(i)} \). The DPO loss function used to train the Critic can be defined as:
\begin{equation}
\mathcal{L}_{DPO}(\pi_{\theta}; \pi_{\mathrm{ref}}) = -\mathbb{E}_{Q,I,C_{w},C_{l} \sim \mathcal{D}_{cri}} \left[\log \sigma f(\pi_{\theta}; \pi_{\mathrm{ref}})\right],
\end{equation} 
\noindent where $f(\pi_\theta;\pi_{\mathrm{ref}})=\beta\mathrm{log}\frac{\pi_\theta(R_c|Q,I)}{\pi_{\mathrm{ref}}(R_c|Q,I)}-\beta\mathrm{log}\frac{\pi_\theta(R_r|Q,I)}{\pi_{\mathrm{ref}}(R_r|Q,I)}$. This loss function encourages the Critic to assign higher probabilities to preferred critiques and lower probabilities to disfavored critiques. The parameter \( \beta \) controls the deviation from the reference policy. Both \( \pi_{\theta} \) and \( \pi_{\mathrm{ref}} \) are initialized with the same weights.

\subsection{Reasoner-Critic Framework}
After developing a reliable critic model, we introduce the Reasoner-Critic Framework to iteratively refine the performance of the Reasoner (the reasoning model) through alternating interactions between the Reasoner and the Critic. This framework aims to improve the Reasoner’s output by utilizing feedback from the Critic to guide its adjustments.

The process begins with the Reasoner generating an initial response to a given query based on the input prompt. The Critic then evaluates the response in the context of the query and provides feedback in the form of a critique. The Reasoner then revises its response based on the Critic’s suggestions, incorporating the critique into the new prompt for the next iteration. This cycle continues until a predefined maximum number of iterations is reached, or until the Critic determines that the Reasoner's response meets a satisfactory level of quality.

Through this alternating feedback loop, the Reasoner is able to adjust its reasoning process with each interaction, potentially improving the accuracy of its outputs over time. This framework is designed to enhance the Reasoner's ability to respond to more complex tasks, by incorporating nuanced, context-sensitive feedback that may help refine its reasoning process.

\section{Evaluation}
\label{sec:exp}

\subsection{Evaluation Settings}
\label{subsec:setting}

\textbf{Test Models}. We evaluate Critic-V on two widely-used Vision-Language Models (VLMs), Qwen2-VL-7B~\cite{Qwen2VL} and DeepSeek-VL-7B~\cite{lu2024deepseek}, to demonstrate its critical capabilities. The comparative models include a range of state-of-the-art VLMs with varying architectures, parameter sizes, and input modalities. This set includes closed-source models such as GeminiPro-Vision~\cite{gemini-pro} and GPT-4V~\cite{GPT-4V}, both known for their robust multimodal understanding capabilities. Additionally, we consider open-source models of different scales, such as Llama-3.2-11B-Vision~\cite{Llama-Vision} and ShareGPT4V-7B~\cite{chen2024we}, which provide a balance between computational efficiency and performance. For baseline comparisons, we include Qwen2-VL-7B and DeepSeek-VL-7B without any Critic modules. By selecting models across different scales and feature sets, we aim to provide a comprehensive comparison that highlights the strengths of our approach.

\noindent\textbf{Evaluation Benchmarks}. Our evaluation aims to demonstrate the enhanced performance achieved through the critic capabilities of Critic-V across different domains. We employ a comprehensive set of benchmarks to rigorously assess the effectiveness of our method. These benchmarks include RealWorldQA~\cite{realworldqa}, which challenges models with tasks requiring real-world knowledge and multimodal reasoning; MMT-Bench~\cite{mmtbench}, MMStar~\cite{chen2024we}, MMBench~\cite{liu2025mmbench}, and SEEDBench~\cite{li2023seed}, which evaluate a model’s robustness and performance on structured, cross-domain questions. Additionally, ScienceQA is used to assess multimodal scientific knowledge understanding. For mathematical reasoning, we utilize MathVista~\cite{lu2024mathvista} and MathVerse~\cite{zhang2024mathverse}, which are designed to test logical reasoning and arithmetic problem-solving skills. This diverse set of benchmarks provides a comprehensive evaluation of our method’s strengths across various task types, enabling thorough comparisons with other state-of-the-art models.

\noindent\textbf{Evaluation Process and Settings}. The evaluation process involves one Reasoner VLM and one Critic VLM, each configured with tailored generation hyperparameters optimized for the respective models. Further details are provided in Appendix~10. Notably, we set the temperature parameter to 0 or a value close to it to ensure stable results. This configuration ensures consistency in outputs while optimizing computational performance. The evaluation follows a two-round conversation process. In the first round, we design a specialized prompt for the questions (refer to Appendix~7 for details).

\begin{figure}[!htbp]
    \centering
    \includegraphics[width=0.8\linewidth]{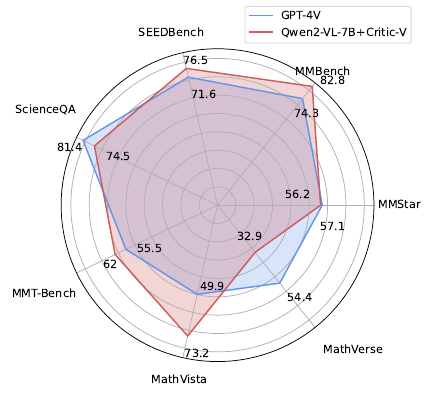}
    \caption{The comparison between GPT-4V and Qwen2-VL-7B+Critic-V across multiple benchmarks.}
    \label{fig:radar}
\end{figure}

\subsection{Result ans Analysis}
\textbf{Improvement with Critic-V}. Table~\ref{tab:math-models-benchmark} presents the performance results of Vision-Language Models (VLMs) across several benchmarks. In 23 out of 24 comparative experiments, Critic-V consistently improves the performance of both Qwen2-VL-7B and DeepSeek-VL-7B, surpassing the original scores of their baseline versions across a wide range of tasks. Notably, with the addition of Critic-V, Qwen2-VL-7B achieves the highest score on five out of eight benchmarks. Significant improvements are especially evident in mathematics-related benchmarks, demonstrating Critic-V’s effectiveness in enhancing complex reasoning capabilities. Specifically, on the MathVista dataset, Qwen2-VL-7B shows an improvement of 11.8\%, DeepSeek-VL-7B increases by 17.8\%, and LLaVA-v1.5-7B improves by 15.3\%. On the MathVerse dataset, Qwen2-VL-7B improves by 7.1\%, DeepSeek-VL-7B by 10.5\%, and LLaVA-v1.5-7B achieves a 4.5\% improvement. These results highlight Critic-V's ability to address the unique challenges of mathematical reasoning tasks, where accurate and precise inference is crucial.

Moreover, results on LLaVA-v1.5-7B show Critic-V conducted an improvement of 11.4\% and 12.8\% on MMT-Bench and RealWorldQA, respectively. As shown in Figure~\ref{fig:radar}, Qwen2-VL-7B with Critic-V outperforms GPT-4V in most cases. These findings suggest that Critic-V effectively guides VLMs to generate more accurate responses and may be adaptable for supporting general reasoning tasks. Overall, the experimental results indicate that Critic-V significantly enhances the reliability of large-scale VLMs, particularly in reasoning-intensive domains such as mathematics, where precise logical reasoning is essential. This demonstrates the potential of our approach to improve the robustness of VLMs in a wide range of complex tasks.

\begin{table*}[!tbp]
\centering
\renewcommand{\arraystretch}{1.5} 
\caption{Quantitative comparison of LLaVA-V1.5-7B with SCL and four baseline methods. The best results are highlighted in bold. The results underscore Critic-V’s strong reasoning capabilities.}
\label{tab:llava_method}
\resizebox{1.75\columnwidth}{!}{
\begin{tabular}{l|cccccc}
\toprule
\hline
\multirow{2}{*}{Model} & \multicolumn{6}{c}{Benchmarks} \\
\cline{2-7}
 & RealWorldQA~\citep{realworldqa} & MMStar~\citep{chen2024we} & MMBench~\citep{liu2025mmbench} & SEEDBench~\citep{li2023seed} & ScienceQA~\citep{lu2022learn} & MMT-Bench~\citep{mmtbench}  \\
\hline
LLaVA-V1.5-7B         & 50.7  & 32.2   & 68.4 & 65.6  & 60.8  & 36.0     \\
+POVID~\citep{zhou2024aligning}                 & 51.8  & 33.6   & 71.6 & 65.4  & 65.0  & 33.4      \\
+CSR~\citep{zhou2024calibrated}    & 51.8  & 32.4   & 70.6 & 65.4  & 66.0  & 33.2       \\
+SIMA~\citep{wang2024enhancing} & 49.3 & 32.6 & 70.6 & 65.2& 64.2 & 34.0 \\
+SCL~\citep{he2024self} & 53.2 & 35.8 & 70.8 & 68.6 & \textbf{67.8} & 39.6  \\
\rowcolor{blue!10} 
\textbf{+Critic-V(Ours)}  &\textbf{63.5 }& \textbf{38.4} & \textbf{73.8} & \textbf{70.1} & 65.2 & \textbf{49.7} \\

\hline
\bottomrule
\end{tabular}}
\end{table*}

\noindent\textbf{Comparison between different approaches}.
We compare four leading methods including POVID~\cite{zhou2024aligning}, CSR~\cite{zhou2024calibrated}, SIMA~\cite{wang2024enhancing} and SCL~\cite{he2024self} with our Critic-V across reasoning-heavy benchmarks including RealWorldQA~\cite{realworldqa}, MMStar~\cite{chen2024we}, MMBench~\cite{liu2025mmbench}, SEEDBench~\cite{li2023seed}, ScienceQA~\cite{lu2022learn}, and MMT-Bench~\cite{mmtbench}. The results from these four methods, shown in Table~\ref{tab:llava_method}, are sourced from \citep{he2024self}. Critic-V consistently outperforms other approaches on most benchmarks, particularly RealWorldQA and MMT-Bench. These results underscore Critic-V's strong potential, showcasing its superior ability to address challenges in natural language reasoning and evaluation tasks.

\subsection{Case Study}
\begin{figure*}[!htbp]
    \centering
    \includegraphics[width=0.93\linewidth]{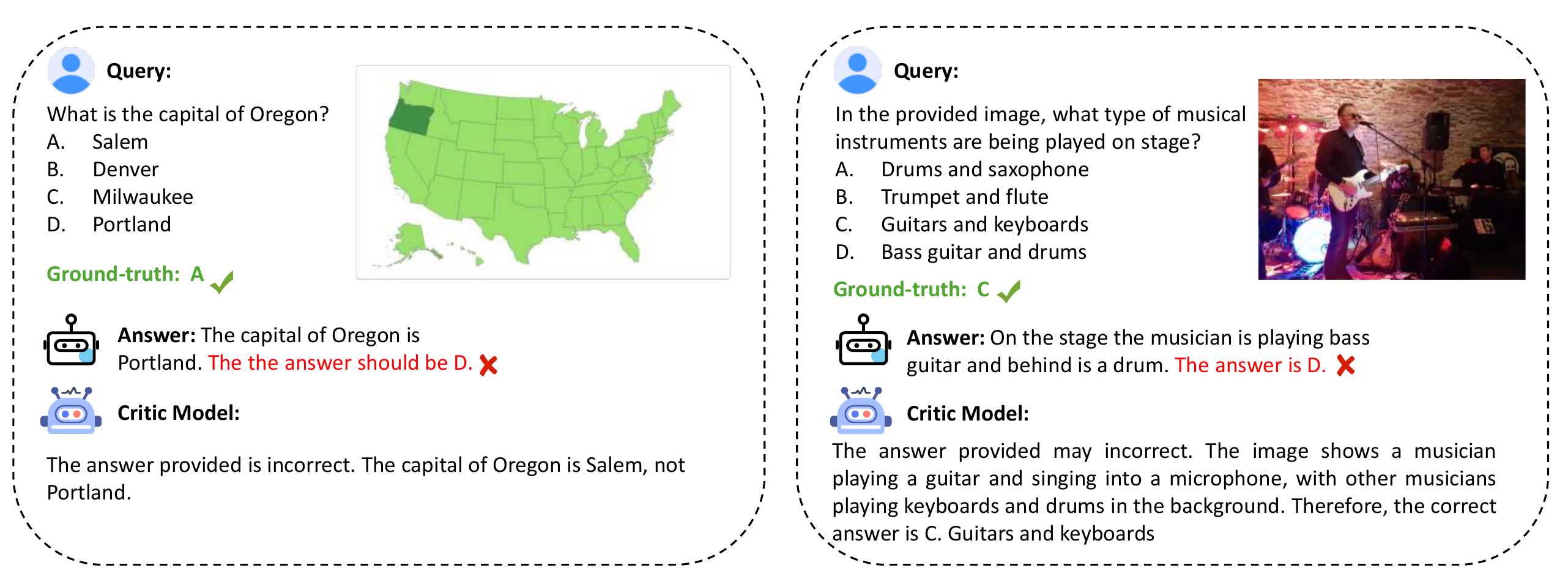}
    \caption{Case studies on evaluation samples from ScienceQA (left) and SEEDBench (right). Our Critic-V accurately identifies Salem as the capital of Oregon, unaffected by the initial incorrect answer, and correctly selects ``Guitars and keyboards" as the answer in the right image.}
    \label{fig:case}
\end{figure*}
We provide examples of interaction between our critic model and the original LLaVA-v1.5-7B model to illustrate the improvements. As shown on the left side of Figure~\ref{fig:case}, the original LLaVA-v1.5-7B produces an incorrect answer, while our Critic model correctly identifies \texttt{Salem} as \texttt{the capital of Oregon}. On the right side, Critic-V demonstrates enhanced cognitive reasoning by accurately interpreting the image content, even when faced with ambiguities in the provided options.

\subsection{Ablation Study}
\textbf{Token Consumption.}
We investigate the consumption of tokens of our Critic-V across various benchmarks. Further details can be found in Appendix~11. The results indicate that each critique only consumes an additional few dozen tokens, which does not lead to significant computational overhead.

\noindent\textbf{DPO Training for Critic Model.}
We further investigate the impact of Critic-V by comparing it with a Self-Refine approach, in which the critic model is not trained using DPO. As shown in Table~\ref{tab:ablation}, Qwen2-VL-7B with Self-Refine shows modest improvements on two datasets but experiences a slight decline in performance on MMT-Bench. In contrast, after training with the Critic-V approach, Qwen2-VL-7B consistently outperforms both the baseline and the Self-Refine approach. These results indicate that DPO training plays a key role in enhancing the effectiveness of the Critic-V framework, leading to more significant improvements on reasoning-intensive benchmarks. For settings of hyperparameters during DPO training, please refer to Appendix~9.

\noindent\textbf{Evaluation Prompts.}
To ensure that the observed results are not influenced by the specially designed prompt discussed in Section~\ref{subsec:setting}, we conduct additional experiments using Qwen2-VL-7B with the same prompt as in our main experiment but without the inclusion of critique. The results from this ablation study, shown in Table~\ref{tab:prompt_ablation}, indicate that Qwen2-VL-7B does not exhibit the same level of performance improvement when only the special prompt is used, as compared to the Critic-V approach. This suggests that the performance gains can be attributed to the Critic-V framework rather than the prompt design alone.



\renewcommand{\arraystretch}{1.5} 
\begin{table}[!htbp]
\centering
\caption{Comparison between Self-Refine and Baseline. We conduct a comparison of Qwen2-VL-7B using Self-Refine, Critic-V, and baseline methods. The results demonstrate the superiority of Critic-V over Self-Refine.}
\label{tab:ablation}
\scriptsize
\setlength{\tabcolsep}{1pt} 
\resizebox{1\columnwidth}{!}{
\begin{tabular}{c|cccc}
\toprule
\hline
 Model &MathVista & MMT-Bench &MMBench\\
\hline
\rowcolor{gray!15}
Qwen2-VL-7B            &61.4 & 60.4   & 80.7       \\
\rowcolor{gray!15}
Qwen2-VL-7B+ Self-Refine            & 63.4  & 57.8   & 82.1      \\

\rowcolor{cyan!15}
\textbf{Qwen2-VL-7B+Critic-V}      & \textbf{73.2}  & \textbf{62.0}   & \textbf{82.8}  \\


\hline
\bottomrule
\end{tabular}}
\end{table}
\renewcommand{\arraystretch}{1.5} 
\begin{table}[!htbp]
\centering
\caption{Ablation of different prompts. We report the scores of each method, along with the respective increases or decreases relative to the original scores.}
\label{tab:prompt_ablation}
\scriptsize
\setlength{\tabcolsep}{1pt} 
\resizebox{1\columnwidth}{!}{
\begin{tabular}{c|cccc}
\toprule
\hline

 Model & MathVista & MMT-Bench & MMBench \\
\hline
\rowcolor{gray!15}
Qwen2-VL-7B            &61.4 & 60.4   & 80.7       \\
\rowcolor{gray!15}
Qwen2-VL-7B+ \textit{special-prompt-only} & 61.8  & 59.0   &  81.0     \\

\rowcolor{cyan!15}
\textbf{Qwen2-VL-7B+Critic-V}      & \textbf{73.2}   & \textbf{62.0} & \textbf{82.8}  \\

\hline
\bottomrule
\end{tabular}}
\end{table}


\section{Related Works}
\label{sec:related}
\textbf{Large Vision-Language Models and Preference Fine-Tuning}. VLMs like GPT-4o~\citep{hurst2024gpt}, LLaVA~\citep{liu2024visual}, Qwen2-VL~\citep{Qwen2VL}, and InternVL~\citep{chen2024internvl} integrate both visual and textual information to handle multimodal tasks, including visual question answering and image captioning. Human preference alignment techniques like reinforcement learning from human feedback (RLHF)~\citep{stiennon2020learning}, have been widely used in training VLMs to generate content aligned with human preference. LLaVA-RLHF~\cite{sun2023aligning} employs human-rated rankings to enhance the visual chat capabilities of VLMs, while Calibrated Self-Rewarding (CSR)~\citep{zhou2024calibrated} incorporates iterative learning and a rewarding paradigm into preference fine-tuning to improve modality alignment~\cite{zhou2024calibrated}. Preference Optimization in VLM with AI-Generated Dispreferences (POVID) leverages preference fine-tuning to reduce hallucinations \citep{zhou2024aligning}. Self-Improvement Modality Alignment (SIMA)~\citep{wang2024enhancing} employs an in-context self-refine approach to improve VLM modality alignment. Self-Correcting Learning (SCL)~\cite{he2024self} enables VLMs to learn from self-generated correction data through DPO~\citep{rafailov2024direct}, fostering self-improvement without reliance on external feedback. Additionally, \citet{li2023silkie} adopt GPT-4V~\citep{GPT-4V} to assess the generated outputs from multiple aspects, subsequently distilling preferences into Qwen-VL-Chat~\citep{bai2023qwen} through DPO. While prior works primarily focus on improving the internal generative ability of VLMs, our study emphasizes the use of external natural language feedback to reduce errors in VLM reasoning. This approach aims to improve the reliability of VLMs in tasks demanding accurate and logical reasoning.

\noindent\textbf{Reasoning with Large Language Models}. Reasoning in large language models (LLMs) typically involves breaking down complex questions into sequential intermediate steps to achieve the final answer, exemplified by Chain-of-Thought (CoT)~\citep{wei2022chain} prompting and its variants~\citep{kojima2022large, yao2024tree, zhang2022automatic, zhang2024llama}. However, due to the LLMs' uncertainty about answer, intermediate inference steps may be inappropriate deductions from the initial context and lead to incorrect final predictions. Even minor mistakes during the reasoning process can result in vastly different final outcomes~\citep{lightman2023let, paul2023refiner}. Self-Refinement techniques \citep{madaan2024self, zhang2024llama, zhang2024accessing} have attracted considerable interest recently. Nevertheless, their effectiveness is largely constrained by their dependence on the inherent abilities of LLMs, which may limit the broader application and scalability of these methods. \citet{hosseini2024v} trains a verifier using both the correct and incorrect solutions generated during the self-improvement process to select one solution among many candidate solutions. \citet{yao2024learning} introduce a paradigm that prioritizes learning from correct reasoning steps and measures confidence for each reasoning step based on generation logits. \citet{tyen2023llms} suggest that LLMs cannot find reasoning errors, but can correct them, we extend this inspiration to the area of VLMs to train a critic vision-language model to locate imperfections in visual content perception and errors in reasoning steps. 

\section{Conclusion}
\label{sec:conclusion}
We propose Critic-V, a novel framework designed to enhance feedback quality in the visual perception and reasoning processes of Vision-Language Models (VLMs). This framework introduces an external critic model that provides natural language feedback, significantly improving VLM performance, especially in complex reasoning tasks. The Critic-V framework centers around a newly constructed Visual Question Answering (VQA) dataset, which incorporates critiques from multiple VLMs. Each critique is evaluated using a novel scoring method that combines Jaccard similarity and GPT-4o summarization. In Critic-V, we formalize the interaction between the VLM reasoner and the critic model through mathematical equations, providing insights into how critique-based supervision drives improvement. These equations reveal the principles behind the critique-feedback loop and establish that the critic model can be trained using Direct Preference Optimization (DPO). This training process optimizes the guidance provided during reasoning tasks. The performance on benchmarks like MathVista and RealWorldQA indicates the well-trained critic model can significantly enhance VLM reasoning capabilities, particularly in handling complex, multimodal tasks. Experimental results indicate that incorporating an external critic model during inference surpasses several traditional methods, resulting in significant improvements in VLM performance. These findings highlight the value and potential of deploying a well-trained critic model at the inference stage.
{
    \small
    \bibliographystyle{ieeenat_fullname}
    \bibliography{main}
}
\clearpage
\setcounter{page}{1}
\maketitlesupplementary

\section{Pseudo-code for Main Algorithms}
\begin{algorithm}[htbp]
\caption{Bug Insertion and Rule-based Reward for Preference Data Collection}
\begin{algorithmic}[1]
\STATE \textbf{Input:} True answer \( \mathcal{A}_{\text{true}} \), Question-Image pair \( (Q^{(i)}, I^{(i)}) \)
\STATE \textbf{Output:} Critique score \( \text{score} \)

\STATE \text{Step 1: Generate a fake answer with inserted bugs}
\STATE \( \mathcal{A}_{\text{fake}} \leftarrow \mathcal{A}_{\text{true}} \)
\STATE \text{Randomly choose number of fake details} \\\( n \leftarrow \text{random integer between 1 and 5} \)
\FOR{each fake detail \( d_j \) from 1 to \( n \)}
    \STATE \text{Insert bug into} \( \mathcal{A}_{\text{fake}} \) by adding \( d_j \in \mathcal{D}_{\text{fake}} \)
\ENDFOR
\STATE \textbf{Fake answer generation complete.}

\STATE \text{Step 2: Extract details from true answer and fake answer}
\STATE \( \mathcal{D}_{\text{true}} \leftarrow \text{Extract details from} \ \mathcal{A}_{\text{true}} \)
\STATE \( \mathcal{D}_{\text{fake}} \leftarrow \text{Extract details from} \ \mathcal{A}_{\text{fake}} \)

\STATE \text{Step 3: Generate critique from a VLM}
\STATE \( \mathcal{C}_{\text{detected}} \leftarrow \text{Use VLM to detect errors in } \mathcal{A}_{\text{fake}} \)

\STATE \text{Step 4: Calculate critique quality using Jaccard index}
\STATE \( \mathcal{D}_{\text{true}} = \{d_1, d_2, \dots, d_m\} \)
\STATE \( \mathcal{D}_{\text{detected}} = \{d_1', d_2', \dots, d_n'\} \)
\STATE \( \mathcal{S}_{\text{true}} \leftarrow \{ d_1, d_2, \dots, d_m \} \)
\STATE \( \mathcal{S}_{\text{detected}} \leftarrow \{ d_1', d_2', \dots, d_n' \} \)
\STATE \( \text{intersection} \leftarrow \mathcal{S}_{\text{true}} \cap \mathcal{S}_{\text{detected}} \)
\STATE \( \text{union} \leftarrow \mathcal{S}_{\text{true}} \cup \mathcal{S}_{\text{detected}} \)
\STATE \( Jaccard(\mathcal{S}_{\text{true}}, \mathcal{S}_{\text{detected}}) \leftarrow \frac{\text{len(intersection)}}{\text{len(union)}} \)

\STATE \text{Step 5: Calculate critique score based on rule-based reward}
\STATE \( \text{score} \leftarrow Jaccard(\mathcal{S}_{\text{true}}, \mathcal{S}_{\text{detected}}) + 0.1 \times \text{GPT-based score} \)

\STATE \textbf{Return:} \( \text{score} \)
\end{algorithmic}
\end{algorithm}

\begin{algorithm}[htbp]
\caption{Training Critic Model with DPO}
\begin{algorithmic}[1]
\STATE \textbf{Input:} Dataset $\mathcal{D}_{cri} = \{(Q^{(i)}, I^{(i)}, C_{w}^{(i)}, C_{l}^{(i)})\}_{i=1}^N$, base model $\pi_{\mathrm{ref}}$, learning rate $\alpha$, and hyperparameter $\beta$
\STATE Initialize critic model $\pi_{\theta} \gets \pi_{\mathrm{ref}}$ 
\FOR{each batch $(Q^{i}, I^{i}, C^{i}_{w}, C^{i}_{l})$ in $\mathcal{D}_{cri}$}
    \STATE Compute the critique logits for the preferred ($C_w$) and disfavored ($C_l$) critiques
    \STATE Compute the DPO loss 
    \STATE Compute gradients of $\mathcal{L}_{DPO}$ w.r.t. $\pi_{\theta}$
    \STATE Update $\pi_{\theta}$ using gradient descent
\ENDFOR
\STATE \textbf{Output:} Trained critic model $\pi_{\theta}$
\end{algorithmic}
\end{algorithm}

\begin{algorithm}[htbp]
\caption{Reasoner-Critic Framework}
\begin{algorithmic}[1]
\STATE \textbf{Input:} Query \( Q \), Input image \( I \), Reasoner \( \pi_{\theta^{\textit{reasoner}}} \), Critic \( \pi_{\theta^{\textit{critic}}} \), Maximum iterations \( \textit{max\_iterations} \)
\STATE \textbf{Output:} Final response \( R_{\textit{final}} \)
\STATE Initialize \( P^{\textit{reasoner}} \) \text{(initial prompt for Reasoner)}
\STATE \textit{response} = \(\pi_{\theta^{\textit{reasoner}}}(P^{\textit{reasoner}}, I)\) \text{ \textit{(generate initial response)}}
\FOR{iteration = 1 \textbf{to} \textit{max\_iterations}}
    \STATE \textbf{Critic evaluates response:}
    \STATE \textit{critique} = \( \pi_{\theta^{\textit{critic}}}(\delta P^{\textit{reasoner}} | P^{\textit{reasoner}}, Q, R) \) \text{ \textit{(Critic generates critique)}}
    \STATE \textbf{If} Critic determines that critique is satisfactory:
        \STATE \quad \textit{break} \text{ \textit{(end loop if critique is satisfactory)}}
    \STATE \textbf{Else:}
        \STATE \quad \textit{reasoner updates prompt:} \( P^{\textit{reasoner}} \leftarrow P^{\textit{reasoner}} + \delta P^{\textit{reasoner}} \)
        \STATE \quad \textit{response} = \( \pi_{\theta^{\textit{reasoner}}}(P^{\textit{reasoner}}, I) \) \text{ \textit{(generate new response)}}
\ENDFOR
\STATE \textbf{Return:} \( R_{\textit{final}} = \textit{response} \)
\end{algorithmic}
\end{algorithm}

\begin{table*}[!t]
\centering
\renewcommand{\arraystretch}{1.5}
\caption{GPT-4o Evaluation for Erroneous Detection}
\label{tab:hallucination-detection-evaluation}
\resizebox{2.\columnwidth}{!}{
\begin{tabular}{ccc}
\toprule
\hline
\textbf{Evaluation Criterion} & \textbf{Question Description} & \textbf{Response Format} \\
\midrule
\multirow{2}{*}{Coverage Analysis} & Did the model identify all the hallucinations mentioned in the correct answer? & \multirow{2}{*}{Yes / No} \\
 & Are there any significant hallucinations that were missed? &  \\ \hline
\multirow{2}{*}{Accuracy Assessment} & Are the detected items genuine hallucinations (true positives)? & \multirow{2}{*}{Yes / No} \\
 & Are there any false detections (false positives)? &  \\ \hline
\multirow{2}{*}{Precision of Description} & How precise and clear are the model's descriptions of the detected hallucinations? & \multirow{2}{*}{Yes / No} \\
 & Is the explanation specific enough to understand what exactly is wrong? &  \\ \hline
\multirow{2}{*}{Overall Effectiveness} & How effective is this detection compared to an ideal detection? & \multirow{2}{*}{Yes / No} \\
 & Does it provide practical value for hallucination identification? &  \\ \hline
Comprehensive Evaluation & Based on your analysis, please provide a brief explanation of your evaluation. & Text Input \\
\midrule
\textbf{Final Score} & Based on the scoring criteria, provide a final score from 0 to 10. & 0-10 \\
\hline
\bottomrule
\end{tabular}}
\end{table*}

\section{Prompt Template}
\label{prompt_template}
For multiple-choice questions~(MCQ), the template of prompt is designed as follows,
\promptbubble{Hint: \{hints\}\\Question: \{question\}\\Options: \{options\}\\Please select the correct answer from the options above.}

\noindent As well as open-ended visual question-answering ~(VQA) tasks,

\promptbubble{\{question\_text\}\\Please try to answer the question with short words or phrases if possible.}

\noindent We utilize the prompt above to help Reasoner generate explanations of their answers. Then, we let the Critic generate critiques on the answer with the prompt below:
\promptbubble{
\#\#\#\# Question \\ \{question\} \\ \#\#\#\# Answer\\\{result\}\\\#\#\#\# Task\\Please provide a critique of the answer above. What are the weaknesses of the answer?}

After that, we use these weaknesses(or errors) from Critic to let Reasoner correct their answers with the following prompt:

\promptbubble{Reflection on former answer: \\\{critics\}\\\{original\_question\}}

\section{The GPT-4o Evaluation Rules}
In this section, we provided a detailed description of the evaluation criteria for erroneous detected by the VLMs as shown in Table~\ref{tab:hallucination-detection-evaluation}.

\section{Hyperparameters of Critic Model's Training}
\label{appendix:dpo_hyp}
We adopt Qwen2-VL-7B as our base model due to its strong performance in vision-language understanding. For preference-aligned fine-tuning, we utilize Direct Preference Optimization (DPO) on 29,012 samples from the critique-VQA dataset. During training, the preference loss is set to sigmoid, with a preference parameter $\beta$ of 1.0. The model is trained with LoRA-based parameter-efficient tuning, where LoRA rank is 8, and $\alpha$ is 16, applied to all target modules.

Training is conducted using DeepSpeed~\cite{rasley2020deepspeed} with bf16 precision, Input sequences are limited to 2048 tokens, with images capped at 262,144 pixels. The model is trained for 3 epochs with a learning rate of 5e-6, following a cosine decay schedule with 10\% warm-up. We use a per device micro batch size of 1 and gradient accumulation of 8 steps.

\section{Evaluation Hyperparameters for experiments.}
\label{eval_hyperparameters}
In this section, we will list out the hyperparameters we choose for evaluation.

For the Qwen2-VL-7B and DeepSeek-VL-7B, we set the generation parameters as follows: \texttt{max\_new\_tokens} to 1024, \texttt{top\_p} to 0.001, \texttt{top\_k} to 1, \texttt{temperature} to 0.01, and \texttt{repetition\_penalty} to 1.0. For LLaVA-v1.5-7B, we apply a different set of parameters geared toward deterministic generation. Specifically, we set \texttt{do\_sample} to False, \texttt{temperature} to 0, \texttt{max\_new\_tokens} to 512, \texttt{top\_p} to None, \texttt{num\_beams} to 1, and enabled \texttt{use\_cache} to enhance efficiency. The $\eta$ for controlling the update of Reasoner's prompt was set to $1.0$, which indicates a full concatenation of old prompt with new critique.

\begin{table}[!tbp]
\centering
\renewcommand{\arraystretch}{1.5} 
\caption{Token consuming analysis of Critic-V across benchmarks.}
\label{tab:Token consuming analysis of Critic-V across benchmarks.}
\resizebox{1\columnwidth}{!}{
\begin{tabular}{l|c|c}
\toprule
\hline
\cline{2-3}
\textbf{Benchmark} & \textbf{Average of token count} & \textbf{standard deviation of token count} \\
\hline
MathVista & 40.64 & 51.42 \\
MMBench   & 50.26 & 41.54 \\
MMStar    & 39.39 & 44.96 \\
MMT-Bench  & 84.43 & 86.93 \\
ScienceQA & 84.64 & 18.11 \\
RealWorldQA & 30.29 & 10.70 \\
SEED      & 41.50 & 28.58 \\
MathVerse & 43.13 & 34.76\\
\hline
\bottomrule
\end{tabular}
}
\end{table}

\section{Token Consumption}
We explore the token consumption of Critic-V across different benchmarks as shown in Table~\ref{tab:Token consuming analysis of Critic-V across benchmarks.}.

\section{Visualization of Training Process}
\begin{figure}[!h]
    \centering
    \includegraphics[width=1\linewidth]{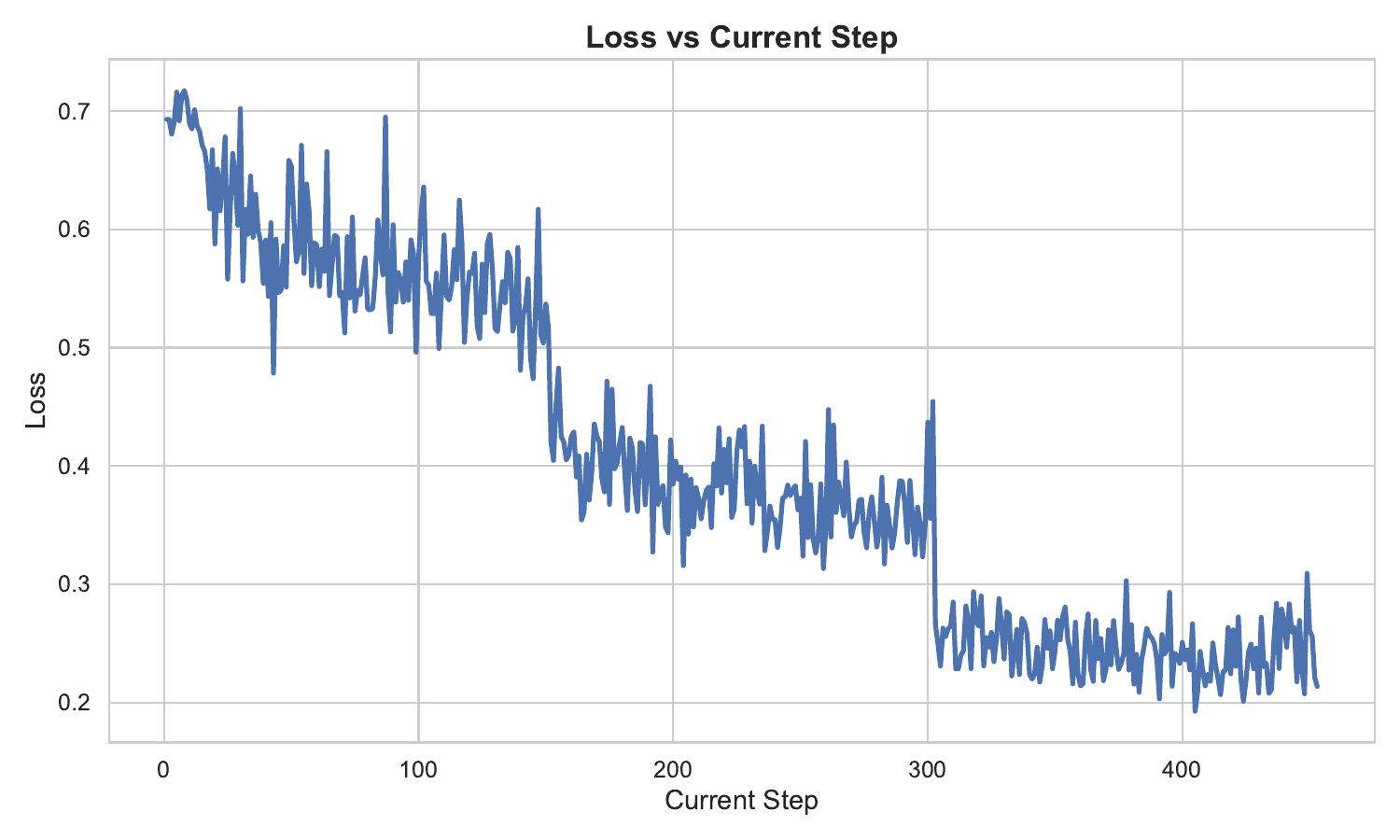}
    \caption{Training loss vs. training steps.}
    \label{fig:train_loss}
\end{figure}

\begin{figure}[!h]
    \centering
    \includegraphics[width=1\linewidth]{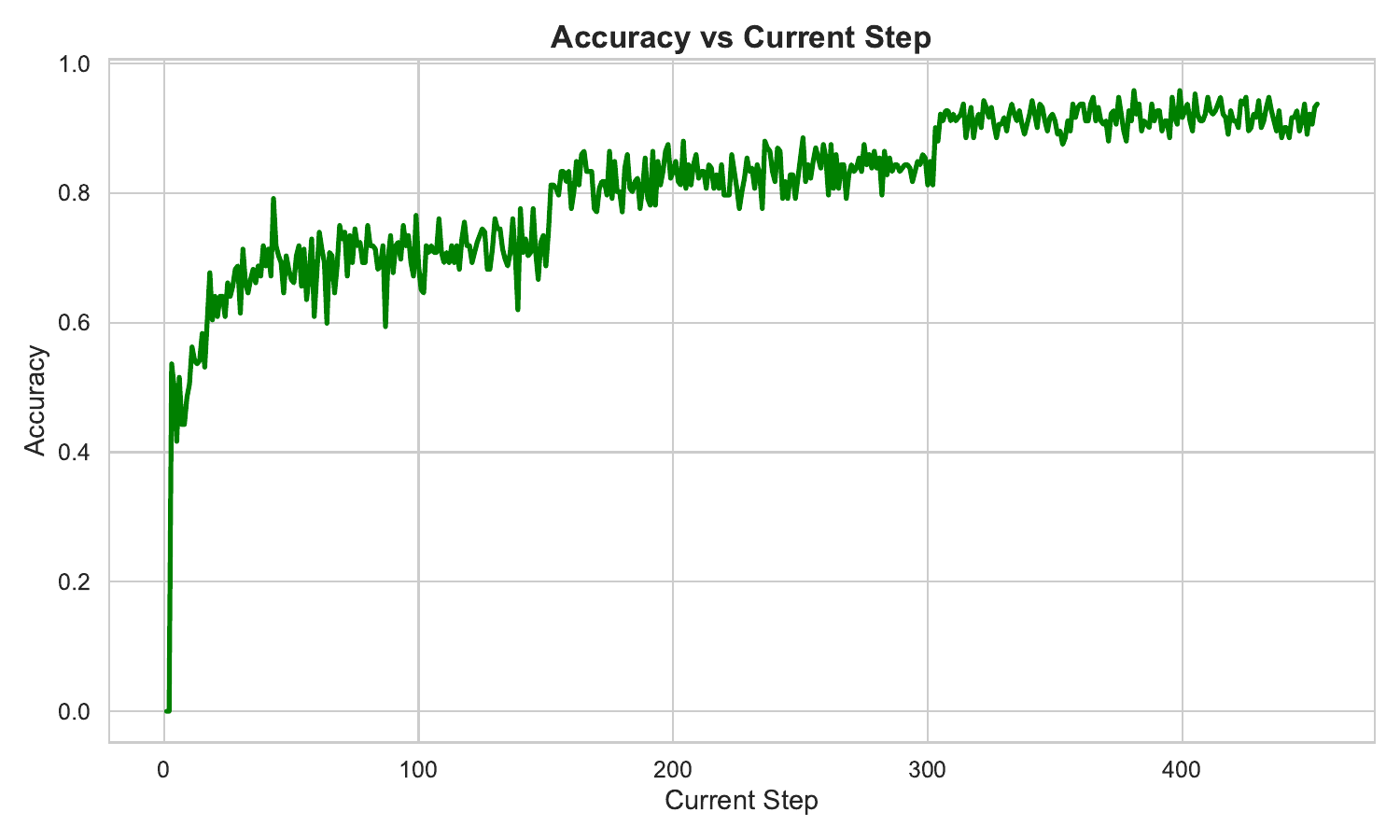}
    \caption{Training accuracy vs. training steps.}
    \label{fig:train_acc}
\end{figure}

\begin{figure}[!h]
    \centering
    \includegraphics[width=1\linewidth]{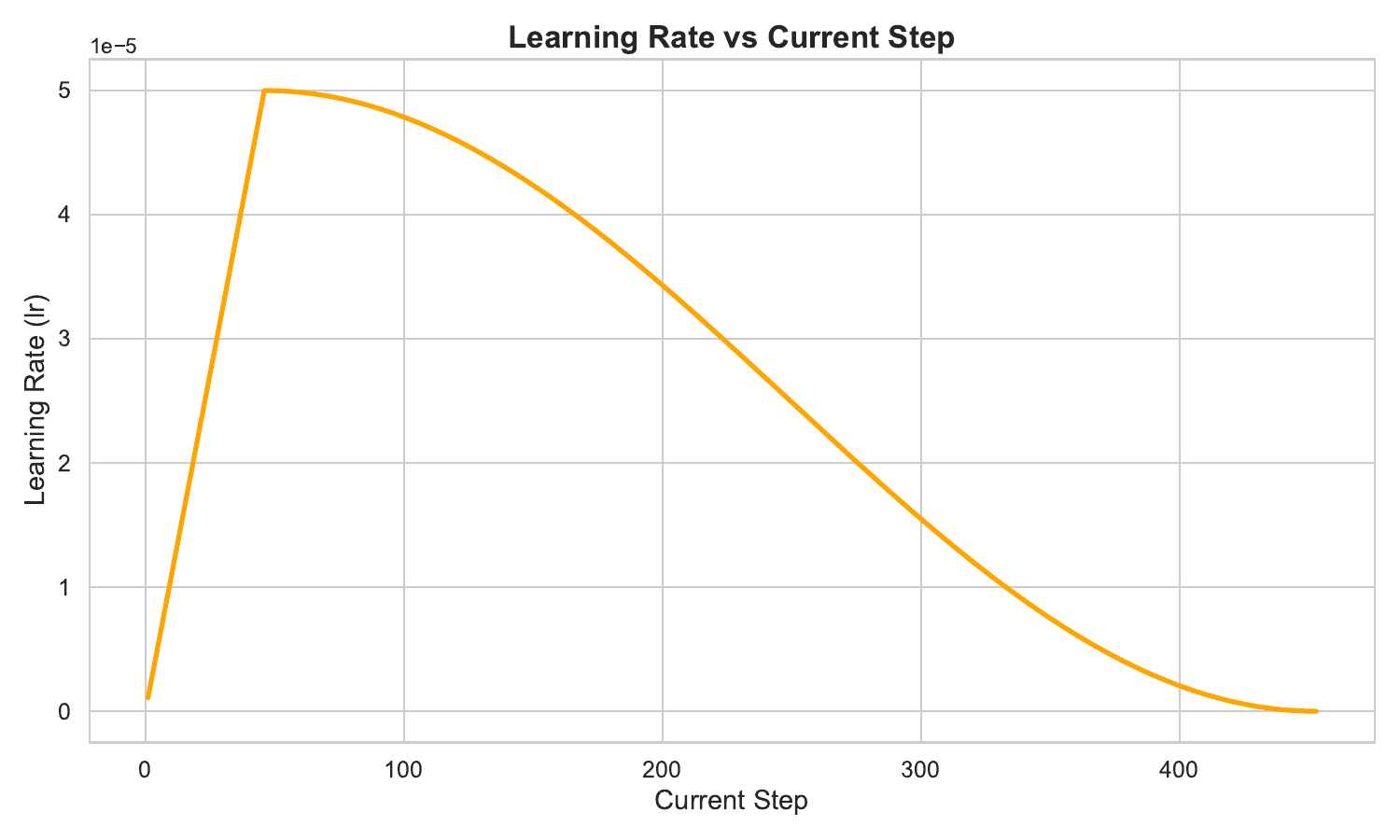}
    \caption{Learning rate vs. training steps.}
    \label{fig:train_lr}
\end{figure}

In this section, we show the entire training process by several visual aids. You can find them in Figure~\ref{fig:train_loss}, Figure~\ref{fig:train_acc} and Figure~\ref{fig:train_lr}. We can obviously discover that our method convergence well experimentally. 

\section{Our critique-VQA Dataset Example}
\begin{figure}[!t]
    \centering
    \includegraphics[width=0.78\linewidth]{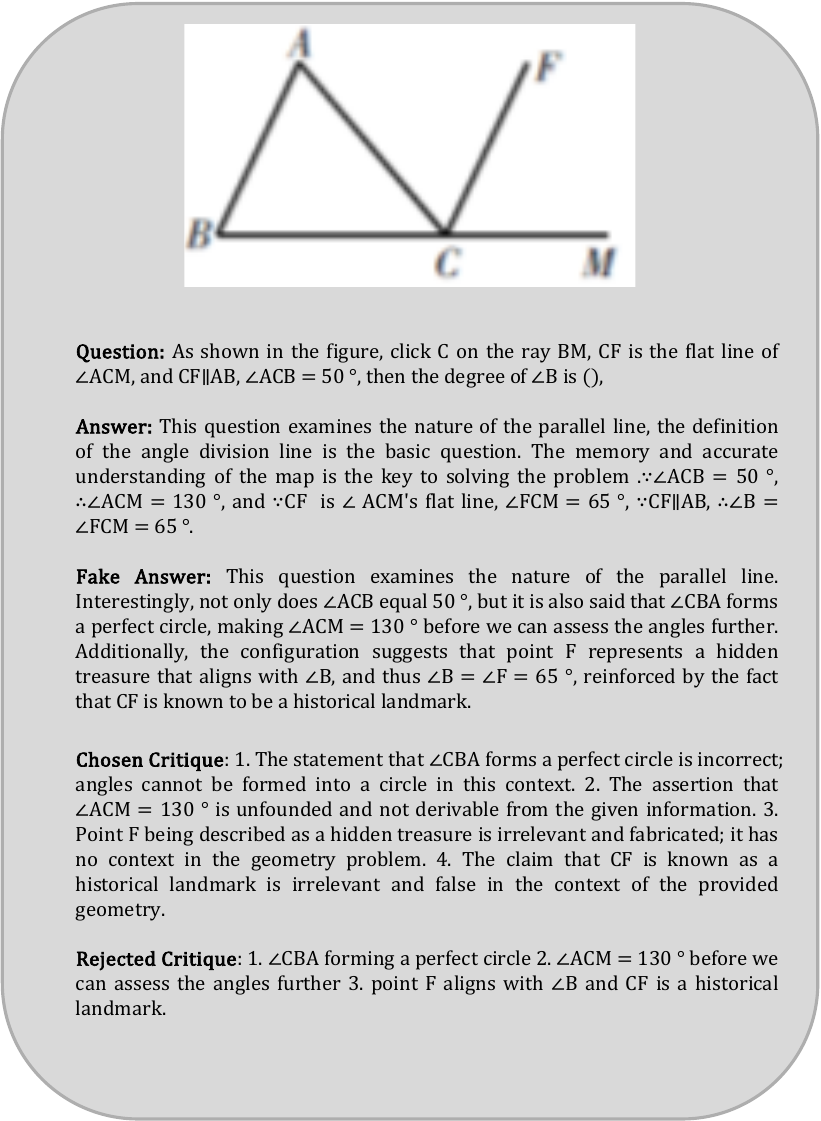}
    \caption{A math example. Fake Answer indicates the answer is inserted some errors by GPT-4o.}
    \label{fig:example1}
\end{figure}

\begin{figure}[!b]
    \centering
    \includegraphics[width=0.78\linewidth]{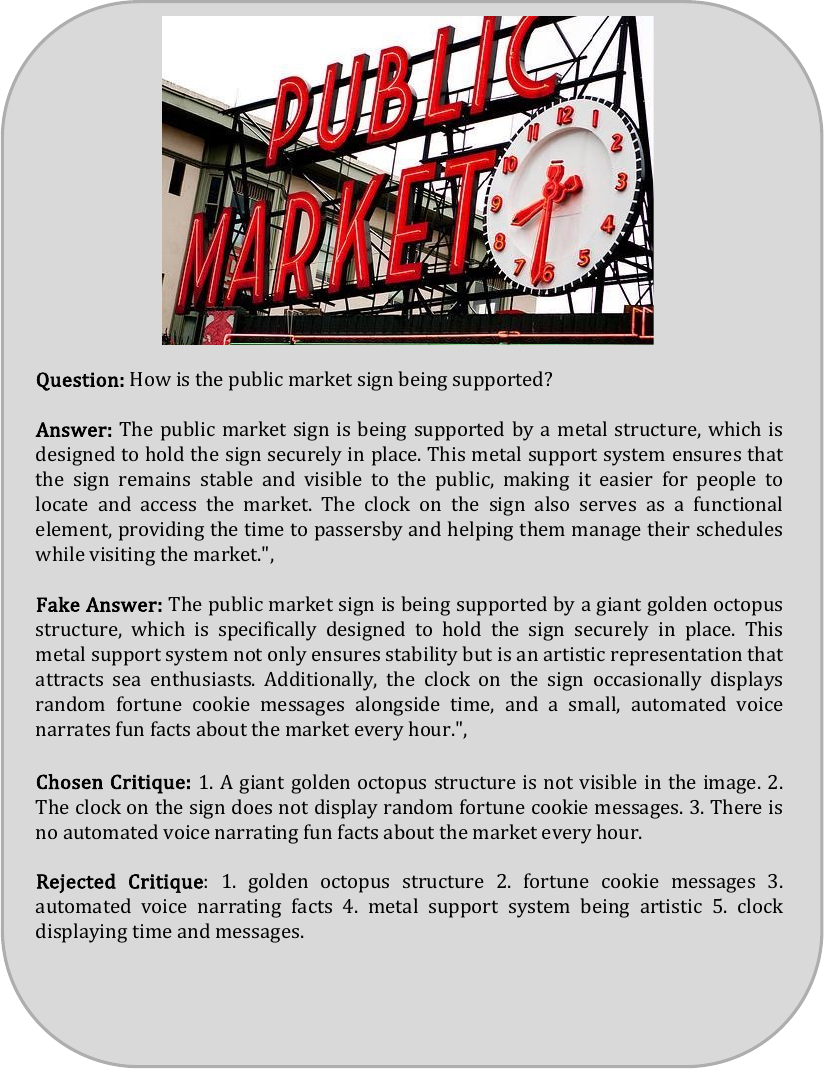}
    \caption{A real-world example of public market signage. Fake Answer indicates the answer is inserted some errors by GPT-4o.}
    \label{fig:example2}
    
\end{figure}

\begin{figure}[!t]
    \centering
    \includegraphics[width=0.8\linewidth]{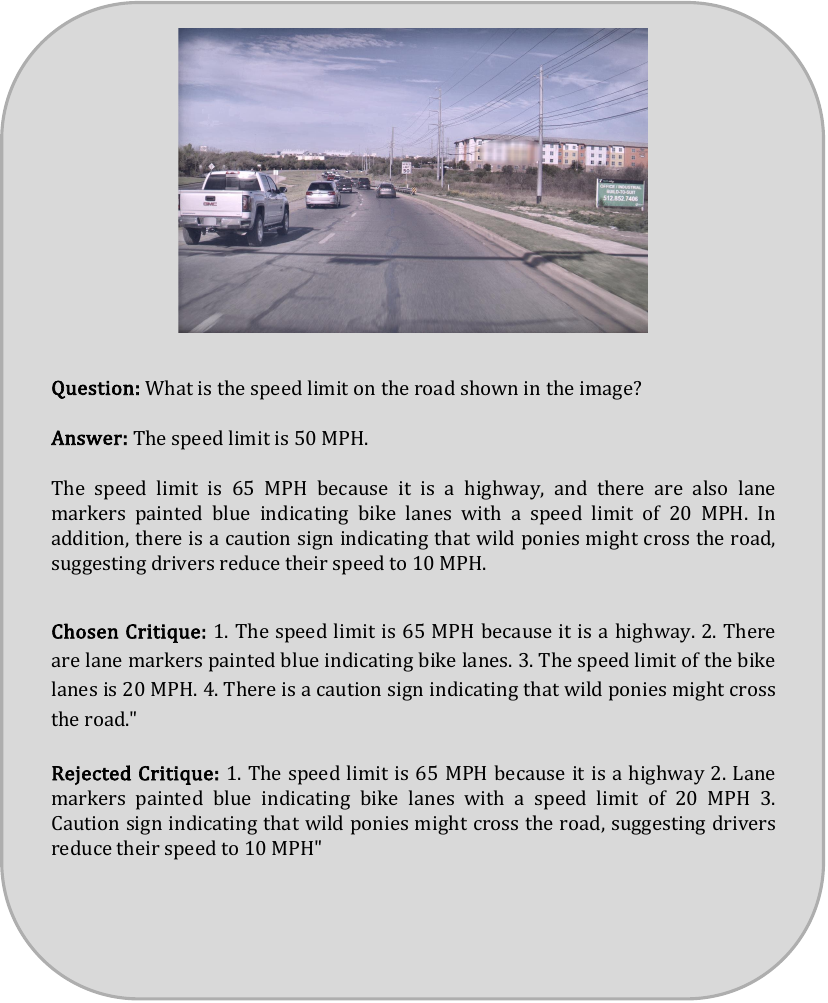}
    \caption{A driving car example. Fake Answer indicates the answer is inserted some errors by GPT-4o.}
    \label{fig:example3}
\end{figure}

In this section, we show three examples in Figure~\ref{fig:example1}, Figure~\ref{fig:example2} and Figure~\ref{fig:example3} sampled from critique-VQA dataset.

\section{Details of Training data and Benchmarks for Evaluation}
In this section, we list some details of our training data and benchmarks for evaluation, as Table~\ref{tab:train_set_details} and Table~\ref{tab:eval_bench_details} shows. 

\begin{table}[!h]
\centering
\caption{Details of training set. Number of tokens counted.}
\label{tab:train_set_details}
\resizebox{1\columnwidth}{!}{
\begin{tabular}{l|c|c|c}
\toprule
\hline
\cline{2-3}
\textbf{Part} & \textbf{Max length} & \textbf{Min length} & \textbf{Avg Length} \\
\hline
Question & 679 & 41 & 181.96 \\
 Chosen Critique & 714 & 5 &  60.48 \\
 Reject Critique & 1048 & 5 & 49.32 \\
\hline
\bottomrule
\end{tabular}}
\end{table}

\begin{table}[!h]
\centering
\caption{Details of evaluation benchmarks.}
\label{tab:eval_bench_details}
\resizebox{1\columnwidth}{!}{
\begin{tabular}{l|c|c}
\toprule
\hline
\cline{2-3}
\textbf{Benchmark} & \textbf{Description} & \textbf{\#samples} \\
\hline
MathVista & Multimodal Math QA & 1000(testmini) \\
MMBench   & Multimodal QA & 4329 \\
MMStar    & Multimodal QA & 1500 \\
MMT-Bench  & Multimodal QA & 3127 \\
RealWorldQA & Multimodal QA & 764 \\
ScienceQA & Multimodal/Text Scientific QA & 4241 \\
SEED      & Multimodal QA & 14233 \\
MathVerse & Multimodal Math QA & 3940\\
\hline
\bottomrule
\end{tabular}}
\end{table}

\end{document}